\documentclass[conference]{IEEEtran}
\IEEEoverridecommandlockouts

\usepackage{marvosym}
\usepackage{verbatim}
\usepackage{cite}
\usepackage{makecell}
\usepackage{amsmath, amssymb, amsfonts}
\usepackage{subfigure}
\usepackage{graphicx}
\usepackage{textcomp}
\usepackage{xcolor}
\usepackage{multirow}
\usepackage{caption}
\usepackage{diagbox}
\usepackage{url}
\usepackage{float}
\usepackage{fancyhdr}
\usepackage{threeparttable}
\usepackage[noend]{algpseudocode}
\usepackage{algorithmicx, algorithm}
\usepackage{epstopdf}
\usepackage{xargs}
\usepackage{geometry} 
\def\BibTeX{{\rm B\kern-.05em{\sc i\kern-.025em b}\kern-.08em
    T\kern-.1667em\lower.7ex\hbox{E}\kern-.125emX}}
    
\newcommandx{\unsure}[2][1=]{\todo[linecolor=red,backgroundcolor=red!25,bordercolor=red,#1]{#2}}
\newcommandx{\improvement}[2][1=]{\todo[linecolor=blue,backgroundcolor=blue!25,bordercolor=blue,#1]{#2}}

\title{Perspective, Survey and Trends: Public Driving Datasets and Toolsets for Autonomous Driving Virtual Test\\

\thanks{This work is by supported by the National Key
R\&D Program of China under Grant  No. 2017YFB0503400, the National Science Foundation of China under Grant No. 11701545, the fund
of the State Key Laboratory of Software Development Environment
under Grant No. SKLSDE-2020ZX-15.

Duing to the page limitations, we have not released detailed information on the toolsets and datasets in this paper. For specific open source datasets and toolsets, please contact ruanli@buaa.edu.cn.}}

\author{
\IEEEauthorblockN {Pengliang Ji, Li Ruan\textsuperscript{\Letter}, and Limin Xiao}
\IEEEauthorblockA{\textit{
State Key Laboratory of  Software Development Environment} \\
\textit{School of Computer Science and Engineering}\\  Beihang University\\ Beijing, China \\{jpl1723@buaa.edu.cn, ruanli@buaa.edu.cn, liminxiao@buaa.edu.cn}}
 
\and
\IEEEauthorblockN{Yunzhi Xue, Qian Dong}
\IEEEauthorblockA{\textit{
State Key Laboratory of Computer Science} \\
\textit{Institute of Software \& Chinese Academy of Sciences}\\
Beijing, China \\
{yunzhi@iscas.ac.cn, dongqian@iscas.ac.cn}}

}

\begin{document}

\maketitle 

\begin{abstract}

Owing to the merits of early safety and reliability guarantee, autonomous driving virtual testing has recently gains increasing attention compared with closed-loop testing in real scenarios. Although the availability and quality of autonomous driving datasets and toolsets are the premise to diagnose the autonomous driving system bottlenecks and improve the system performance, due to the diversity and privacy of the datasets and toolsets, collecting and featuring the perspective and quality of them become not only time-consuming but also increasingly challenging. This paper first proposes a Systematic Literature review approach for Autonomous driving tests (SLA), then presents an overview of existing publicly available datasets and toolsets from 2000 to 2020. Quantitative findings with the scenarios concerned, perspectives and trend inferences and suggestions with 35 automated driving test tool sets and 70 test data sets are also presented. To the best of our knowledge, we are the first to perform such recent empirical survey on both the datasets and toolsets using a SLA based survey approach. Our multifaceted analyses and new findings not only reveal insights that we believe are useful for system designers, practitioners and users, but also can promote more researches on a systematic survey analysis in autonomous driving surveys on dataset and toolsets.

\end{abstract}
\section{Introduction}
Safety is the most critical requirement in autonomous  driving test at the era when the automation level of autonomous driving cars reaches SAE3 or higher where  the driving responsibility is gradually shifting from traditional drivers to autonomous driving cars\cite{safe-example1} \cite{4142919}. Recent safety research\cite{safe-example1}\cite{safe-example2}\cite{safe-example3} shows that autonomous driving still faces huge challenges in complex interactive scenarios and unstructured environments, such as campuses with surging people and high-speed fog on highways. For example, Uber recorded the first serious fatal accident of an autonomous driving car \cite{accident1}. Extensive research \cite{safe-example2}\cite{safe-example3} shows that autonomous driving faces more demanding safety and reliability requirements than traditional automotive. 

\subsection{Challenges and Problems} 
With the rapid development of deep learning that covers end-to-end learning, 
owing  to  the  merits  of  early  safety  and  reliability guarantee, autonomous driving virtual testing has recently gains increasing  attention  compared  with  closed-loop  testing  in  real scenarios \cite {8324372}\cite{8835477}. With  the rapid development of big data-driven, extensive research  shows the  availability  and  quality  of  autonomous driving  datasets  and  toolsets  are  the  premise  to  diagnose  the autonomous driving system bottlenecks and improve the system performance  \cite{2003DRIVABILITY}\cite{8418447}\cite{9022151}\textcolor{red}. However, due  to  the  diversity  and  privacy  of  the  datasets and toolsets, collecting and featuring the perspective and quality of  datasets and toolsets for virtual testing of autonomous driving become  not  only  time-consuming  but  also  increasingly challenging.

However, the recent research  of autonomous driving datasets and toolsets  in autonomous driving virtual testing faces two main challenges
as follows:  
\begin{itemize}
    \item \textit{How to survey the recent increasing datasets and toolsets for autonomous driving  virtual test using a systematic method?}
 Recently there are some surveys, for example\cite{9090897} and \cite{8760560}, show the urgent demands about the datasets and environments of autonomous driving test. However, they paid more attention to the results analysis and a formal systematic  survey process for autonomous driving test still lacks. 
 \item  \textit{ What are the recent findings and trends of both   toolsets and datasets with scenario concerned so far?}
Although datasets and toolsets are the foundations of autonomous tests and there are some recent survey papers on datasets\cite{CGV-079}\cite{Feng_2020}, only \cite{8667012} recently reviewed the virtual test platforms. However,  \cite{8667012} paid more attention on exploring the simulator of autonomous driving, but lacks tools to establish other testing processes. Moreover, they are based on the papers before 2018 which lacks those of 2020. 
\end{itemize}

\subsection{Main Contributions}
To solve these essential challenges in autonomous driving virtual test, we presents a survey of datasets and toolsets for autonomous driving virtual test from 2000 to 2020. The major contributions of this paper are summarized as follows:
\begin{itemize}
    \item To the best of our knowledge, we are the first  propose a Systematic Literature Review (SLR) approach for autonomous driving tests.
    
    \item We present a review of the datasets and toolsets for automated driving virtual tests based on the investigation research from 2000 to 2020.
   Quantitative  findings  with  the  scenarios concerned,  perspectives  and  trend  inferences  and  suggestions with the most important 35  automated  driving  test  tool  sets  and  70  test  data  sets are also presented. To the best of our knowledge, we are the first to perform such recent empirical systematic survey on both the datasets and toolsets  using  a  SLA  based  survey  approach.
   
\end{itemize}

    
    

\subsection{Organization}
\begin{figure}[!t]
  \centerline{\includegraphics[width=8.68cm,height=4.4cm]{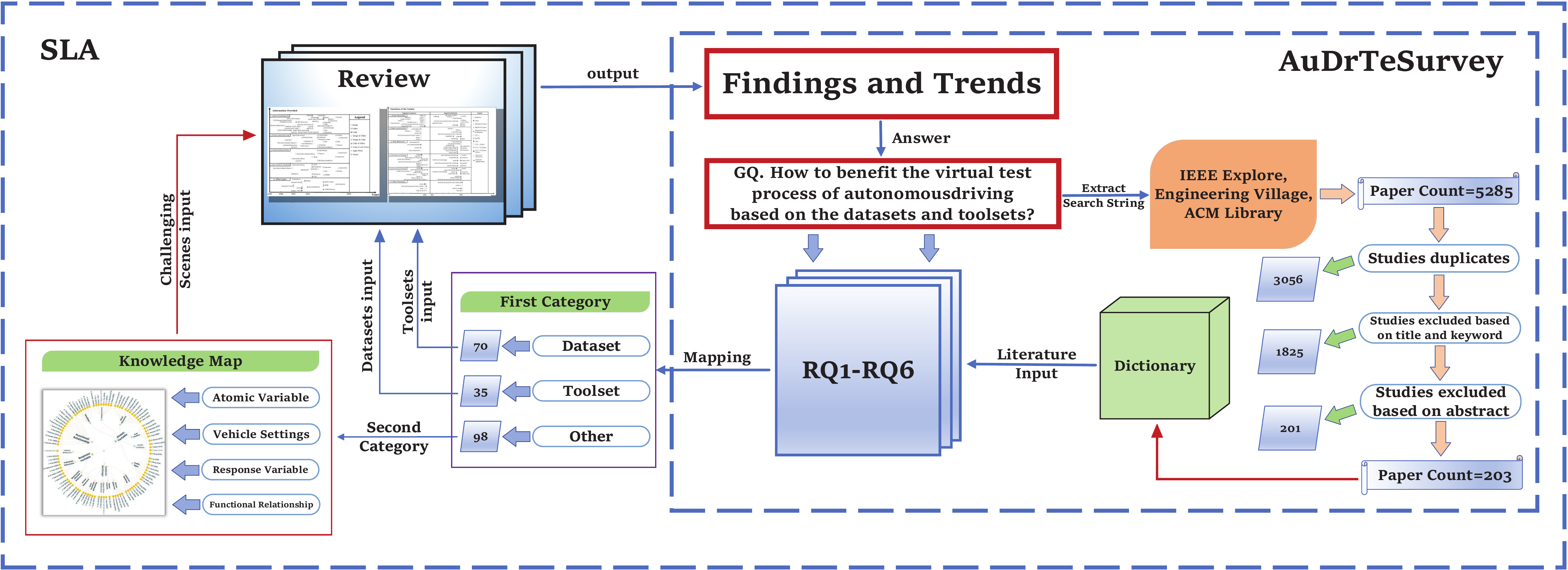}}
  \caption{Autonomous Driving Virtual Test Systematic Literature Review framework.}
  \label{Fig:Framework}
\end{figure}
The remainder of this paper is organized as follows:  Section \ref{sec:Review method} introduces the SLA based autonomous driving research survey approach. Section\ref{sec:Recent Perspective, Trends} introduces results and trends inferences. The paper is concluded in Section\ref{Conclusion}.

\section{AuDrTeSurvey: Systematic Literature Review Approach for Autonomous Driving Virtual Test}\label{sec:Review method}
In this section, we introduce the overall design and process of our systematic literature review approach for autonomous driving virtual test.
\subsection{Overall Design}
As is introduced in \cite{article}, Systematic Literature Review (SLR) is a mainstream means of understanding the latest research tools in a specific field\cite{2006Mobile}. 
We introduce SLA into autonomous driving survey process. Therefore, our Systematic Literature Review (SLR) approach for Autonomous Driving Virtual Test abbreviated as AuDrTeSurvey will provide autonomous driving virtual test specific survey rules to help researchers identify and explain the available research related to collecting datasets and toolsets for autonomous driving virtual test questions. AuDrTeSurvey mainly includes searching, evaluating, explaining, and summarizing findings and trends inference from autonomous driving research. Based on the scientific process, 
AuDrTeSurvey can provide clear motivation for new work, and new evidence to guide decision-making using autonomous driving datasets and toolsets \cite{article2}.

\subsection{Process}

Based on the framework in Fig.\ref{Fig:Framework}, we now present the details of our analysis process. The key steps  are highlighted as follows:
\begin{enumerate}
\item Design research questions in autonomous driving virtual test.
  \item Perform a knowledge map based test influencing factors analysis.
 First, the most critical papers are classified into sub-categories which  will be denoted as Atomic Variable, Response Variable,  Vehicle Settings and the relationship between variables. Then based on the first-level categories, sub-categories (or atomic factors) are derived. After these steps, a knowledge map can be formed.
 \item Perform a review of datasets and toolsets in autonomous driving test. 
   
  \item Analyze new findings and give research trend inferences with the scenes, influence factors concerned.
 \label{Process:findings}
 \end{enumerate}

Using the literature from 2000 to 2020 as an example, the analysis process and its results  will be shown in the following Section\ref{sec:Recent Perspective, Trends}.


\section{Case Study Design}\label{sec:Recent Perspective, Trends}


\subsection{Design of Research Questions}
As our method is driven by research questions, first  the  general question (GQ)  and the sub-questions are designed.
\begin{itemize}
    \item GQ. How to benefit the virtual test process of autonomous driving based on the datasets and toolsets?
\end{itemize}

Next, GQ was divided into the following six research questions (RQ), and we will mark at the answer to them in full text.
\begin{itemize}
    \item RQ1. Research from which perspective has an impact on autonomous driving testing?
    
    \item RQ2. What data sets and tools for automated driving tests are involved in the research?
    
    \item RQ3. Which test scenario variables are mainly involved in the research?
  
    \item RQ4. Which technique was used in the test?
    
    \item RQ5. What innovative discoveries or unique features did the research illustrate?
     
    \item RQ6. What are the unsolved challenges and future trends ?
    
\end{itemize}

\subsection{Knowledge Map based Test Influencing Factors Analysis}

\begin{figure}[!t]
  \centerline{\includegraphics[width=3.45in]{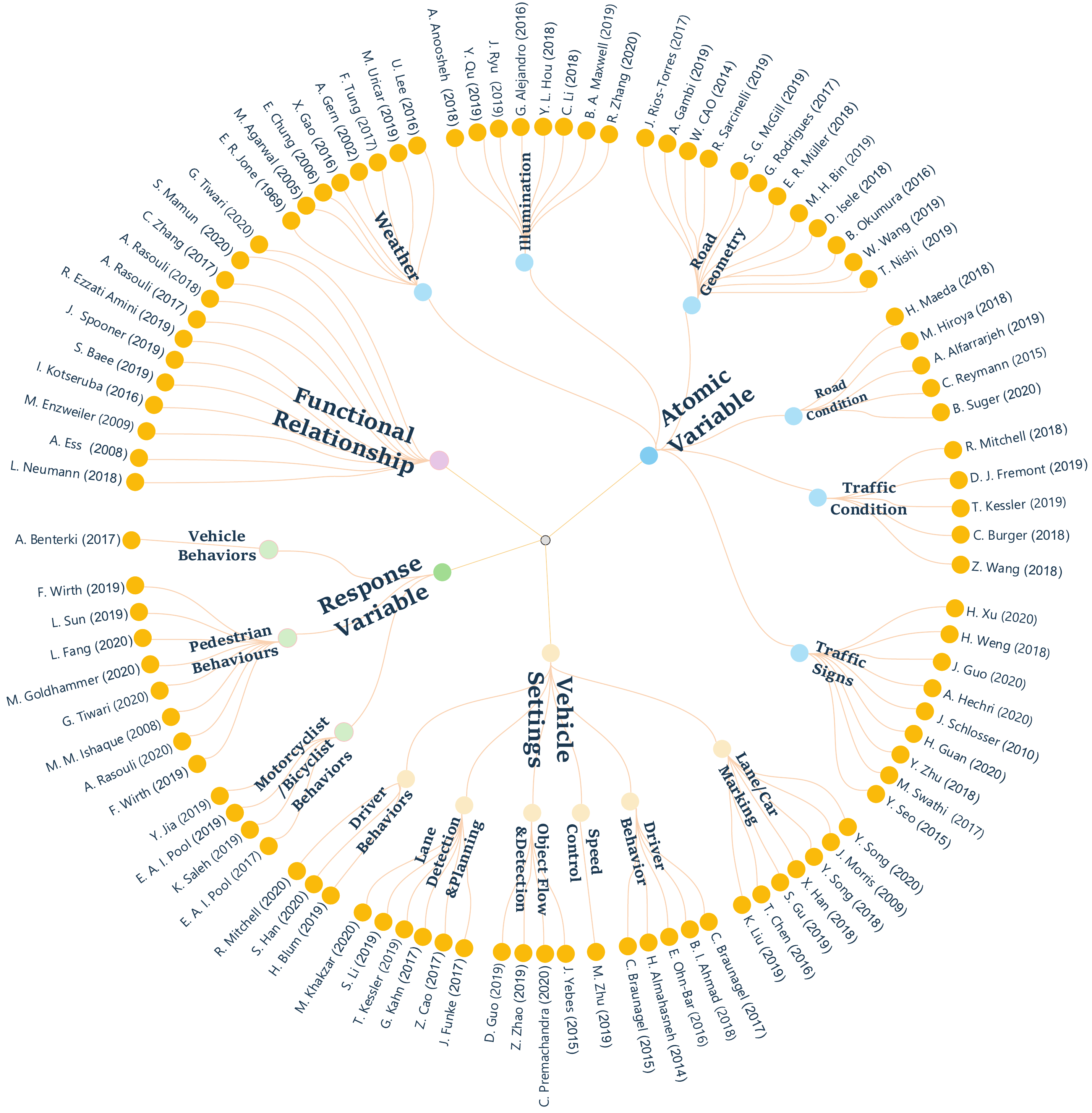}}
  \caption{Knowledge Map based Test Influencing Factors Analysis.}
  \label{fig1}
\end{figure}
\begin{figure*}[htbp]
  \centering
  \subfigure[Datasets]{\label{fig2}\includegraphics[width=3.4in]{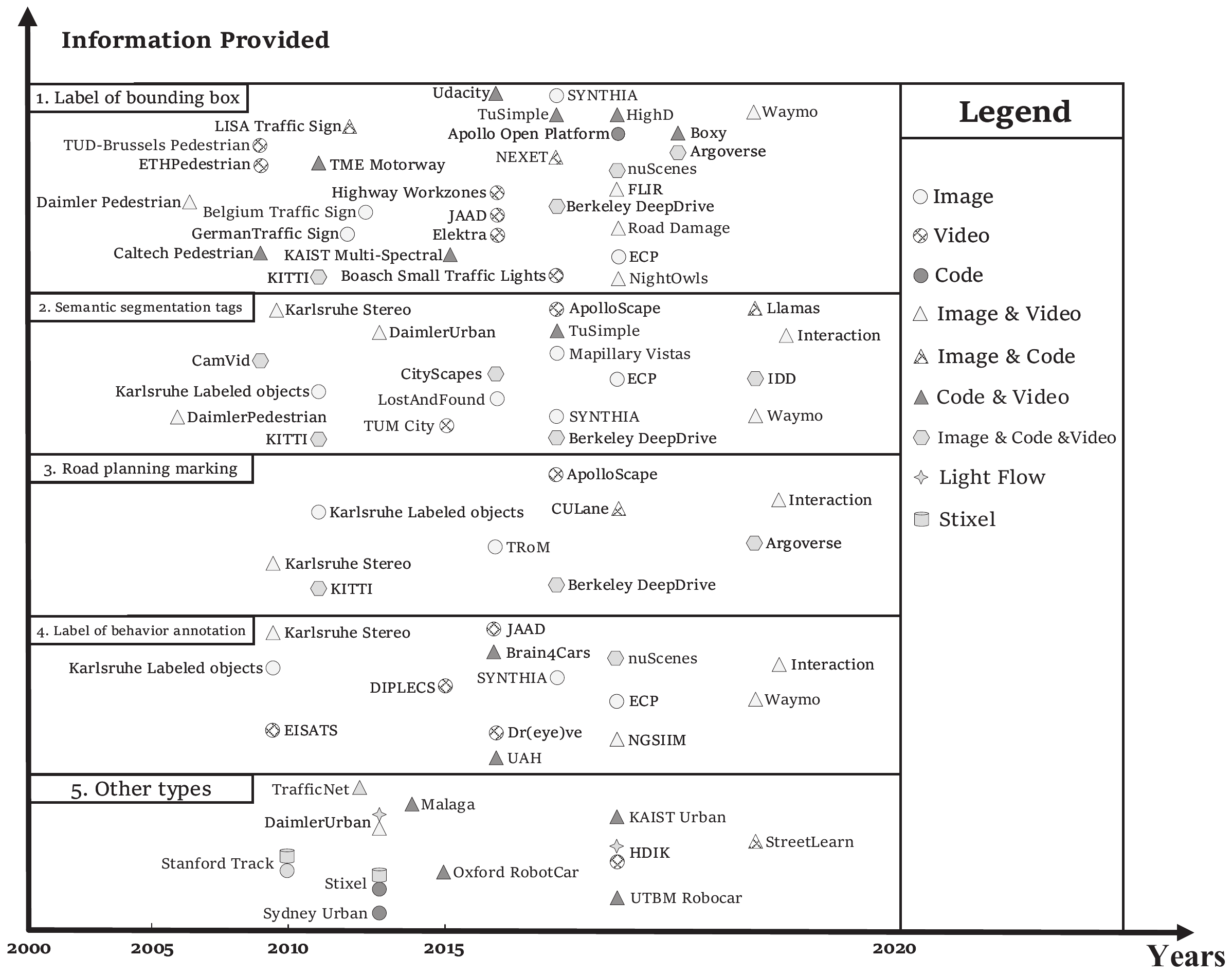}}
  \subfigure[Toolsets]{\label{fig3}\includegraphics[width=3.4in]{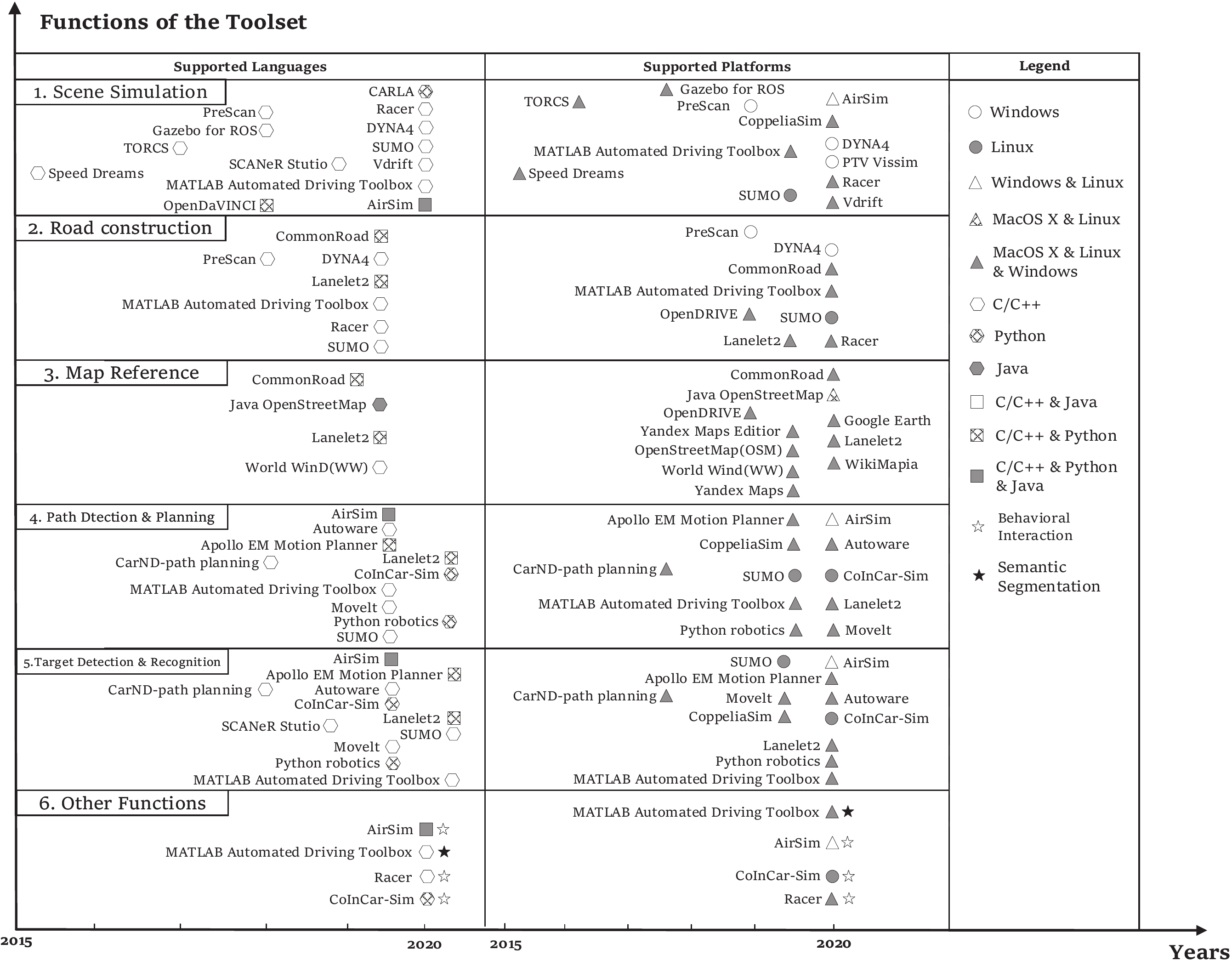}}
  \caption{Scenario Driven Autonomous Driving Datasets and Toolsets Survey}
  \label{fig_datasetsandtoolsets}
\end{figure*}

Next, the latest progress of 98 papers are presented using the knowledge map (Fig.\ref{fig1}). The 98 papers are classified into four categories which include test scene elements denoted as Atomic Variable, interaction under vehicle response denoted as Response Variable, the setting of the algorithm, technology, etc. of the test denoted as Vehicle Settings, the relationship between variables denoted as Functional Relationship to help answer general question (RQ1). Based on the categories, 15 factors are derived, together with its challenging scenes and datasets we recommended are shown in Tab.\ref{Knowledge map}.

\subsection{A Review of Scene Driven Autonomous Datasets and Toolsets}
For datasets, we reviewed 70 datasets, covering basic information, open source links, data set format, size, provided markup information and other data items were taken into consideration (RQ2). Based on our analysis, a summary of autonomous test datasets with scenario and influencing factor concerned is shown in Fig.\ref{fig2}. 

For toolsets, we investigated 35 toolsets and included basic information, open source links, main functions, supported languages and platforms and other data items into the discussion (RQ2). A summary of autonomous test datasets with scenario and influencing factor concerned  is shown in Fig.\ref{fig3}.

\section{Findings and Trends Inferences}\label{sec:Findings and Trends Inferences}
\subsection{Findings and Trend Analysis: Datasets}
\begin{table}
	\caption{Summary of Autonomous Test Datasets With Scenario and Influencing Factor Concerned}
	\label{Knowledge map}
	\setlength{\tabcolsep}{3pt} 
	\renewcommand\arraystretch{1.5}
\begin{tabular}{|m{1.7cm}<{\centering}|m{3.6cm}<{\centering}|m{2.6cm}<{\centering}|}  
\hline
     Factor (RQ3)  & Challenging Scenes  &Typical Datasets\\
    \hline
    \multicolumn{3}{|c|}{Atomic variable}\\
    \hline
    Weather &  Blizzard, Heavy fog, Torrential rain, Overcast clouds & ApolloScape, Dr(eye)ve, FLIR\\
    \hline
    Illumination & Shadow, Night, Direct sunlight, Alternating light and dark & Daimler Urban, NightOwls, SYNTHIA\\
    \hline
    Road geometry & Intersections, Roundabouts, Unmapped areas, Country road  & Berkeley DeepDrive, UTBM Robotcar, Malaga\\
    \hline
    Road condition &Bumpy road surface, Unmarked road, Muddy soil, Mountainous area&Road Damage, Waymo, GIDA\\
    \hline
    Traffic condition &High-traffic,  High-speed, Accident-prone intersections & Daimler Urban, Berkeley DeepDrive, KAIST Urban
 \\
    \hline
    Traffic signs &Different traffic lights in different regions, Road sign, Road director, Fence & Berkeley DeepDrive, German Traffic sign, LISA Traffic Sign
\\
    \hline
    \multicolumn{3}{|c|}{Vehicle Settings}\\
    \hline
    Lane/car marking & Broken and missing markings, Irregular lane and road shapes  & ApolloScape, CULane, Interaction
\\
    \hline
    Driver behaviors & Distraction, Drowsiness  & Dr(eye)ve, JAAD  \\
    \hline
    Speed control & Driving on highways or speed limit section, Sudden braking, Overtaking& Highway Workzones,  KAIST Urban, NightOwls
\\
    \hline
    Opject flow \&detection & Small objects moving at high speed, Flying objects, Suddenly appearing objects  & Brain4Cars, DIPLECS, HighD
 \\
    \hline
    Lane detection \&planning & Irregular Lanes, Crossing multiple lanes, Lanes with traffic rules & CULane, NGSIIM, Argoverse
\\
    \hline
    Semantic segmentation & Difficult to capture or fuzzy mark, Road signs with cultural differences & Karlsruhe Labeled objects,  SYNTHIA, LostAndFound\\
    \hline
    \multicolumn{3}{|c|}{Response Variable}\\
    \hline
    Vehicle behaviors & \multirow{3}{*}{\makecell*[{}{p{3.8cm}<{\centering}}]{Overtaking intention, Sudden movement, Non-compliance role, Walk freely in trod, Whistle, Interactive challenge}}& Brain4Cars, UAH, Oxford RobotCar\\
    \cline{1-1}\cline{3-3}
    Pedestrian behaviours &  &Stanford Track,  JAAD, TUM City
\\
    \cline{1-1}\cline{3-3}
    Motorcyclist/ Bicyclist behaviors & & Daimler Pedestrain, Road Damage, ETH Pedestrain\\
    \hline
    \multicolumn{3}{|c|}{Functional Relationship}\\
    \hline
    \multicolumn{3}{|p{8.1cm}<{\centering}|}{Functional relationship is the Interrelationships between factors. For example, Pedestrians will change their behavior as weather or car speed changes. Because there is no quantitative relationship network, plus the difficulty to build, related tests are often skipped. However, It is generally considered as the future work (RQ6). To date, some incipient studies have been carried out, as is shown in Fig.\ref{fig1}.}\\
    \hline
    \end{tabular}
	\label{tab1}
\end{table}



\subsubsection{Cooperated University and Industry Research Enhanced}

As one of the main trend with the datasets, the datasets  not only differs in scale and volume, but also tend to be relatively heterogeneous. For example, these datasets cover testing tasks in different directions, such as perception, mapping and driving strategies, and targets different urban areas, pedestrian behaviors, and scene sources. On the one hand, in the past, data sets developed by companies or schools in Europe and the United States represented as Germany MPI-IS have dominated the mainstream (59 items, 84.3\% in our case) datasets.

On the other hand, recently, the cooperated university and industry research is enhanced. For example, in our study, data sets in Asia represented by Baidu's Apollo  have gains increasing popularity  which amounts to 11.4\%. 

\begin{table*}[htp]
	\caption{Scenario  Driven Autonomous Driving Datasets Survey }
	\label{table}
	\setlength{\tabcolsep}{3pt} 
	\renewcommand\arraystretch{1.5} 
	\centering
\begin{tabular}{|c|m{2cm}<{\centering}|m{2.4cm}<{\centering}|c|m{2.4cm}<{\centering}|m{2.4cm}<{\centering}|c|m{2.3cm}<{\centering}|m{2.2cm}<{\centering}|}
\hline
Time & Dataset & Volume & Time & Dataset & Volume & Time & Dataset & Volume\\
\hline
2020 & Interaction & multiple vary &2017&Bosch Small Traffic Lights & 13K Image &2013& AMUSE &7 Video, 1.2TB\\
\hline
2020 & NGSIIM & multiple vary&2017&DAVIS& 41 Video, 450GB&2013& Daimler Urban &5K stereo Image\\
\hline
2019&Argoverse& 300K Video&2017& Mapillary Vistas&25K Image&2013&Stixel &2.5K Image, 3GB\\
\hline
2019&Boxy &  200K Image&2017&NEXET&55K Image. 10GB&2013&Sydney Urban &200MB\\
\hline
2019& IDD &10K Image&2017&SYNTHIA& 200K Image&2012&German Traffic Sign&5K Image, 1.6GB\\
\hline
2019& Llamas & 100K Image&2017& TuSimple &5K Video&2012&HCI Challenging Stereo&11 Video\\
\hline
2019& StreetLearn & 114K Image &2016&Brain4Cars&700 Video&2012&LISA Traffic Sign&6.6K Image, 8GB\\
\hline
2019& Waymo & multiple vary &2016&CityScapes &25K Image, 63GB&2012&TrafficNet&10.3GB\\
\hline
2018& Apollo Open Platform &multiple vary&2016& Comma.ai & 80GB &2011&Belgium Traffic Sign&9K Image, 50GB\\
\hline
2018& ApolloScape & 114K Image &2016&Daimler Pedestrian&2.5MB\url{~}45GB each&2011&CMU&16 Video, 275GB\\
\hline
2018 &Comma2k19 & 2019 Video, 100GB&2016&Dr(eye)ve&74Video, 555K Image&2011&Karlsruhe Labeled Objects&1K Image, 631MB\\
\hline
2018&CULane &133K Image &2016&Elektra&9 datasets, 0.5GB\url{~}12GB each&2011&KITTI&multiple vary \\
\hline
2018&DBNet & 10K Image&2016& GIDA &multiple vary&2011&TME Motorway&28 Video\\
\hline
2018& ECP & 43K Image &2016&JAAD &347 Video, 170GB&2010&Cheddar Gorge&329GB Video\\
\hline
2018 & FLIR &14K Image &2016& Lost \& Found &2K Image, 40GB&2010&EISATS&multiple vary \\
\hline
2018& HDIK & 1K Image &2016& TRoM & 700 Image &2010&Karlsruhe Stereo&20 Image sequence\\
\hline
2018& HighD &147h Video&2016& UAH& 35 Video&2010& Stanford Track &14K tracks, 5.7GB\\
\hline
2018& KAIST Urban&19Video, 1GB\url{~}22GB each&2016&Udacity &300GB &2009&Caltech Pedestria&250K Image, 11GB\\
\hline
2018& NightOwls & 279K Image &2015&DIPLECS&4.3GB \& 1.1GB&2009& CamVid&700 Image, 8GB\\
\hline
2018& nuScenes & 1K Video, 1.4M Image&2015& HighWay Workzones&6 Video, 1.2GB&2009& ETH Pedestrian&4.8K Image, 660MB\\
\hline
2018 & Road Damage & 9K Image&2015& KAIST Multi-Spectral& 10 Video&2009&Ford&2datasets, 78GB \& 119GB\\
\hline
2018& TUM City & 73GB \& 1.1GB&2015& Oxford RobotCar &130 Video, 23TB&2009&TUD Brussels Pedestrian&1.6K Image\\
\hline
2018 & UTBM Robotcar & 11 Video& 2014&CCSAD &96K Image, 50GB &  &  & \\
\hline
2017&Berkeley DeepDrive &100K Image, 1.8TB&2014&Malaga &15 Video, 70GB & & & \\
\hline
\end{tabular}
	\label{tab1}
\end{table*}

\subsubsection{Technical Trend Findings}
Under the guidance of Section\ref{Knowledge map}, the contribution of the datasets under three test challenges: key cases, high complexity and coverage, and interactive behavior will be discussed, so as to deal with them getting help from datasets (RQ4, RQ5).

\textbf{Key Cases}: 
     Dealing with challenging scenes shown in Tab\ref{Knowledge map} is the way to construct key cases. For example, among datasets, HCI Challenging Stereo proposed a dynamic object evaluation mask and the most extreme weather videos for the response test of the weather environment. Dr(eye)ve covers the most diverse changing weather used to train an end-to-end deep network. For rugged roads, Road Damage is the first unified road damage data set covering the most instances based on a state-of-the-art object detection method using CNN. CCSAD, CULane, etc. also provide guidance in this area. In addition, Bosch Small Traffic Lights propose a traffic light detector based on deep learning and stereo vision to provide the most comprehensive traffic lights information for multi-vehicle interactive testing.

   \textbf{High Complexity and Coverage}:   
    Urban traffic scenarios are popular choice for testing with high coverage. For example, ApolloScape for urban scenes covers the most complex traffic flow videos of traffic participants based joint self-localization and segmentation algorithm. KAIST Urban covers the most diverse radar and point clouds of important features of urban environments by the SLAM algorithm. Cityscapes exploit large volumes of annotated data composed of thirty elements to train DNN. Besides, Daimler Urban,  Lost\&Found, JAAD, etc. are also one of the best choice under high complexity.
    
    \textbf{Interactive Behavior}:
As the key to the safety of the test\cite{8667866}, interactive test and datasets borrowed more and more pedestrian and driver's studies. ETH modeled the most accurate pedestrian trajectory that can be used to train DNN. Based on annotation data and attribute statistics, JAAD focuses on the most typical factors affecting people's attention. Dr(eye)re is the first  dataset to record the driver's attention based on a multi-branch deep architecture. In addition, DIPLECS, Elektra, UAH, etc. also cover better interaction data.





  \begin{table*}
	\caption{Scenario  Driven Autonomous Driving Toolsets Survey}
	\label{table}
	\setlength{\tabcolsep}{3pt} 
	\renewcommand\arraystretch{1.5} 
	\centering
\begin{tabular}{|c|m{2.5cm}<{\centering}|m{2cm}<{\centering}|c|m{2.5cm}<{\centering}|m{2cm}<{\centering}|c|m{2.5cm}<{\centering}|m{2cm}<{\centering}|}
\hline
Time & Toolset & Provider & Time & Toolset & Provider & Time & Toolset & Provider\\
\hline
2020 & AirSim  & Microsoft Inc. &2020
 & MATLAB AD Toolbox
 & The MathWorks
 & 2020& Yandex Maps Editior
 & Yandex Inc.\\
\hline
2020 & Apollo EM Motion Planner &Baidu Inc.&2020 & Movelt
 &PICKNIK Inc.
 & 2019&  OpenDRIVE
 & ASAM\\
\hline
2020 & Autoware & Tier IV Inc.&2020 & OpenStreetMap
 &OpenStreetMap
 &2019& PreScan
 & SIEMENS\\
 \hline
2020 & CARLA & CVC, Spain&2020 & PTV Vissim
 & PTV Group & 2019 & SCANeR Stutio &AVSimulation Inc.\\
\hline
        2020 & CarMaker
 & IPG Inc.&2020 & Python robotic
 & Atsushi Sakai &2018 & AutoSim&NVIDIA \\
\hline

        2020 & CoInCar-Sim
 & KIT Univ.&2020 & Racer
 & Cruden Inc.&2018 &  CarND-path planning 
 &Jeremy Shannon \\
 \hline
        2020 & CommonRoad
 & TUM Univ.&2020&SUMO &OpenMobility Group&2018 &Gazebo for ROS
 &Open Robotic\ \\
 \hline
        2020 &CoppeliaSim
 & Coppelia Robotics Inc.
 &2020 &VDrift
 & VDrift Group
   & 2017&  ASM Traffic
 &dSpace Inc.\\
  \hline
        2020 & DYNA4
 & Vector Inc.&2020 &VTD& VIRES, Germany
 & 2017& OpenDaVINCI
 &Chalmers\\
 \hline
        2020 & Google Earth
 & Google Inc.
 &2020 & WikiMapia
 & WikiMapia Group
 &2017 &TORCS
 &  SOURCEFORG\\
 \hline
        2020 & Java OpenStreetMap
 & Trac Inc.
 &2020 & World Wind(WW)
 & NASA
 &2015 &  Speed Dreams
 & SOURCEFORG\\
 \hline
        2020 & Lanelet2
 & CVC, Spain
 &2020 & Yandex Map
 & Yandex Inc.
 & & & \\
 \hline\end{tabular}
	\label{tab2}
\end{table*}
\subsection{Findings and Trend Analysis: Toolsets}

\subsubsection{Cooperated University and Industry Research Enhanced}

The combination of most tool sets and test can play a good role. As table \ref{tab2} shows, Numerous tools used for autonomous driving have been released by many companies, such as Google Inc. and university, such as Chalmers Univ.. Although there are not publicly emphasize that these tools can be used for autonomous driving tests, as shown in Fig.\ref{fig2}, these tool have wide range of uses in test.

Nowadays, As shown in Fig.\ref{fig3}, the simulation platform is the most popular function (19 items, 54.3\%). In addition, Path detection (12 items, 34.2\%) and Map-based road network construction (11 items, 31.4\%) also plays an important role. For example, PreScan, Google Earth long-term performed well, Airsim is the rise of this field.

\subsubsection{Technical Trend Findings}
Under the guidance of Section \ref{sec:Review method}, three toolset hot spots: simulated scenes diversity and authenticity, road network generation, and motion planning, which reveal the technical challenges of setting up tests: scenario simulation, road network generation, and trajectory planning, will be be discussed to provide assistance in solving above chanllenge (RQ4, RQ5).

\textbf{Simulated Scenes Diversity and Authenticity}:
High-quality sensors, multivariate variables, real-time environments, and real-time response are the key in this area. For example, CoppeliaSim, PreScan are equipped with the most excellent sensors to respond. SCANeR Stutio covers the Most diverse environment scenes, including terrain, pedestrians, etc.,and DYNA4 has the best effect in the simulation of scenes such as lanes. TORCS, Speed Dreams, and OpenDaVINCI perform best in real-time 
respond. In addition, cross-domain research has also been introduced to the test, such as the drone simulator AirSim, and the game GTA.

\textbf{Road Network Generation}:
    Map, Descriptive low-level tools, and Simulator developed on the underlying file is the key in this area. For example, OpenStreetMap provides the most time-efficient and accurate world map reference. OpenDRIVE is the most efficient and underlying foundation for road network tools, Lanelet2 has the best open-access motion planning solution, and CommonRoad has a set of public road network scenarios covering traffic data. AutoSim analyzed the OpenStreetMap to generate the most simulated 3D road network.  Moreover, DYNA4 and MATLAB Automated Driving also contributes better in this area.

\textbf{Motion Planning}:
Nowadays, Taking traffic rules and the intervention of pedestrian into account is popular in motion planning. For example, CoInCar-Sim provides better vehicle trajectory planning program based on the framework of interaction between traffic participants and traffic rules. Apollo EM Motion Planner provides More efficient trajectory schemes under the framework of safety and scalability. CarMaker designed a better simulation solution and semantic recognition of traffic signs to contributes to this area.

\section{Conclusion}\label{Conclusion}

In this paper, we proposed the first systematic literature review method for autonomous  driving virtual test, and presents a survey of the perspectives and trends of both datasets and toolsets for autonomous driving virtual test.  Quantitative   findings with the scenarios concerned, perspectives and trend inferences and suggestions with 35 automated driving test tool sets and 70 test data sets are also presented.
Our future work includes the wide applications of the datasets and toolsets to our virtual tests.


\bibliographystyle{IEEEtran}
\bibliography{mybib}

\end{document}